\documentclass[sigconf]{acmart-me}

\usepackage{booktabs} 
\usepackage{url}
\usepackage{color}
\usepackage{multirow}
\usepackage{enumitem}
\usepackage{array}
\usepackage{siunitx}
\usepackage[labelfont={bf}]{caption}
\usepackage{balance}
\usepackage{array} 

\setcopyright{rightsretained}

\acmDOI{}

\acmISBN{}

\acmConference[MediaEval'18]{Multimedia Evaluation Workshop}{29-31 October 2018}{Sophia Antipolis, France} 
\acmYear{}
\copyrightyear{}

\acmPrice{}

\begin{document}
\title{The Medico-Task 2018: Disease Detection in the Gastrointestinal Tract using Global Features and Deep Learning}

\author{Vajira Thambawita\textsuperscript{1,3}, Debesh Jha\textsuperscript{1,4}, Michael Riegler\textsuperscript{1,3,5}, P{\aa}l Halvorsen\textsuperscript{1,3,5},\\ Hugo Lewi Hammer\textsuperscript{2}, H{\aa}vard D. Johansen\textsuperscript{4}, and Dag Johansen\textsuperscript{4}}
\affiliation{\textsuperscript{1}Simula Research Laboratory, Norway \ \ \ \  \textsuperscript{2}Oslo Metropolitan University, Norway \ \ \ {}
\textsuperscript{3}Simula Metropolitan, Norway\\
\textsuperscript{4}University of Troms{\o}, Norway \ \ \ \
\textsuperscript{5}University of Oslo, Norway \\ }
\email{Contact: vajira@simula.no, debesh@simula.no}

%
%
%
%
%

\renewcommand{\shortauthors}{Thambawita et. al.}
\renewcommand{\shorttitle}{Medico: The 2018 Multimedia for Medicine Task}

\newcolumntype{P}[1]{>{\centering\arraybackslash}p{#1}}
\newcolumntype{x}[1]{>{\raggedleft\hspace{0pt}}p{#1}}

\begin{abstract}
In this paper, we present our approach for the 2018 Medico Task classifying diseases in the gastrointestinal track. We have proposed a system based on global features and deep neural networks. The best approach combines two neural networks, and the reproducible experimental results signify the efficiency of the proposed model with an accuracy rate of 95.80\%, a precision of 95.87\%, and an F1-score of 95.80\%. 
\vspace{-8pt}
\end{abstract}

%
%
%
%
%

\maketitle
\section{Introduction}
\label{sec:intro}


Our main goal for the Medico Task  \cite{Pogorelov2018} is to classify findings in images from the Gastrointestinal (GI) tract. This task provides two types of input data: Global Features (GFs) and original images. The 2017 Medico Task consisted of a balanced dataset with only 8  classes \cite{Pogorelov:2017:KMI:3083187.3083212} whereas the current task consists of a highly imbalanced dataset with 16 classes \cite{nerthus2017mmsys, Pogorelov:2017:KMI:3083187.3083212}, i.e., making this years task more complicated. Different approaches have been used in the last year medico task \cite{2017_paper_1,2017_paper_2,2017_paper_3,2017_paper_4,2017_paper_5, Mediaeval_2017}  based on GFs extractions and Convolutional Neural Networks (CNN) methods. We extend upon these solutions and present our solutions based on both GFs and transfer learning mechanisms using CNN. We achieve best results combining two CNNs and using an extra multilayer perceptron to combine the outputs of the two networks. 

\section{Approaches}
\label{sec:approach}
We approach the problem of GI tract disease detection with small training datasets using five different methods: two based on GF extractions, and three based on CNN with transfer learning described below. 

\vspace{-8pt}
\subsection{Global-feature-based approaches}

 {\textbf{Method 1} and \textbf{Method 2} use the concept of GFs. For the extraction of GFs, we use Lucence Image Retrieveal (LIRE) \cite{lux2016lire}. GFs are easy and fast to calculate, and can also be used for image comparison, image collection search and distance computing \cite{2017_paper_3}. Based on \cite{riegler2017annotation,pogorelov2017efficient}, we use Joint Composite feature   (JCD), Tamura, Color layout, Edge Histogram, Auto Color Correlogram and Pyramid Histogram of Oriented Gradients (PHOG). These features represent the overall properties of the images. Adding more GFs is possible, but it may increase the redundant information which can reduce the overall classification performance. 
 
 The extracted features are sent to the different machine learning classifier for the multi-class classification.  \textbf{Method 1} makes the use of extracted GFs that are sent to SimpleLogistic (SL) classifier. We input the  same selected set of features to the logistic model tree (LMT) classifier in \textbf{Method 2}.}

\vspace{-10pt}
\subsection{Transfer learning based approaches}

Our CNN approaches use transfer learning mechanism with pre-trained models using the ImageNet dataset \cite{ILSVRC15}. Resnet-152 \cite{DBLP:journals/corr/HeZRS15} and Densenet-161 \cite{DBLP:journals/corr/HuangLW16a} have been selected, and this selection is based on top 1-error and top-5-errors rate of pre-trained networks in the Pytorch \cite{paszke2017automatic} deep learning framework.  

\begin{figure}
\texttt{\includegraphics[scale=1.25]{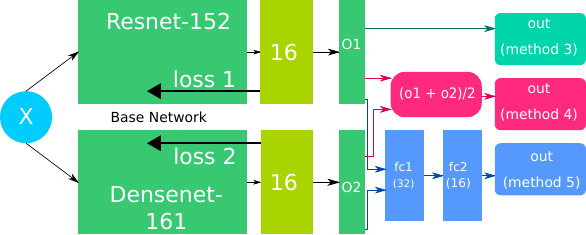}}
\caption{Block diagram of the CNN methods}
\label{fig:models_4_and_5}
\vspace{-10pt}
\end{figure}

 One of the main problems of the given dataset is the "out of patient"-category which has only four images while other classes have a considerable number.  The colour distribution of this class shows a completely different colour domain  compared to the other categories. We identified this difference via manual investigations of the dataset and moved all four images of this category into the corresponding validation set folder. Then, the training set folder is filled with random Google images which are not related to the GI tract. To overcome the problems of stopping training in a local minima, we use the stochastic gradient descent \cite{DBLP:journals/corr/Ruder16}  method with dynamic learning rate scheduling. The losses (loss 1 and loss 2 in Figure \ref{fig:models_4_and_5}) of CNN methods were calculated for each network separately. Additionally, horizontal flips, vertical flips, rotations and re-sizing data augmentations have been applied to overcome the problem of over-fitting. 

\textbf{Method 3} uses transfer learning with Resnet-152 which has the top-1-error and top-5-error rates. The last fully connected layer of Resnet-152, which is originally designed to classify 1000 classes of the ImageNet dataset, has been changed to classify the 16 classes in the MEdico task. Usually, the transfer learning freezes pre-trained layers to avoid back propagation of large errors. This is because of newly added layers with random weights. However, we did not freeze the pre-trained layers, because modifying only the last layer cannot propagate huge errors backwards in transfer learning. The network was trained until it reached to the maximum validation accuracy of the validation dataset.

\textbf{Method 4} extends Method 3 by using two parallel pre-trained models, Resnet-152 and Densenet-161, to get a cumulative decision at the end as depicted in Figure \ref{fig:models_4_and_5}. The  classification is based on an average of the two output probability vectors. Finally, one loss value was calculated and propagated for updating weights.  However, this yields a  restriction of updating weights of networks Resnet-152 and Densenet-161 separately as they required. Therefore, we calculated two different loss values (loss 1 and loss 2 in Figure \ref{fig:models_4_and_5}) from each network to update their weights separately. Both networks were trained simultaneously until it reached to the best validation accuracy by changing hyper-parameters manually.

\textbf{Method 5} was constructed to overcome the limitation of calculating the average of the probabilistic output of the two networks used in Method 4.  Instead of calculating the average using the simple mathematical formula,  another multilayer perceptron (MLP) has been merged with the above network to identify complex mathematical formula to get the cumulative decision as illustrated in Figure \ref{fig:models_4_and_5}.  Therefore, we passed the probability output of two networks (16 probabilities from each network) to a new MLP with 32 inputs, 16 outputs (via sigmoid layer) and one hidden layer with 32 units. In this, we used pre-trained Resnet-152 and Densenet-161 using the dataset and froze them before training the MLP.  Then, we trained only the MLP to identify the best mathematical formula to get the cumulative decision. 

\section{Results and Analysis}

 We have divided the development dataset into a training set (70\%) and a validation set (30\%). For the GFs based approach, ensembles of six extracted GFs were fetched to all the available machine learning classifiers (with different parameters) using WEKA\cite{hall2009weka} library. The SL and LMT classifiers outperform all other available classifiers for the dataset. The other promising classifier were Sequential minimal optimization (RBF kernel), and  a combination of PCA  with LibSVM (RBF) classifier. 
 
On validation set, all the CNN methods (3-5) show accuracies of around 95\% and specificities of around 99\%. These are always better than the GFs based extraction methods (1,2) which have accuracies of around 82\% and specificities of around 98\%. According to the task organizers' evaluation results of the test dataset, Methods 3 to 5 show accuracies and specificities  of around 99\% again,which demonstrates our CNN methods are not overfitted with validation dataset.

Method 5 and 4 with Resnet-152 and Densenet-161 performs better compared to the Method 3 which has only Resnet-152 because of the capability of deciding the final answer based on two answers generated from two deep learning networks. However, getting a cumulative decision based on simple averaging function (Method 4) shows poor performance than the decision taken from a MLP (Method 5). As a result, Method 5 shows better results than method 4 by increasing the accuracy from 0.955 to 0.958. Therefore, Method 5 has been selected as our best method and confusion matrix represented in Table \ref{tab:cm} was generated. An overview of the individual results obtained from five different experiments along with their performance metrics is presented in Table \ref{tab:our_results}. Results obtained from the organizers for the test dataset is presented in the Table \ref{tab:org_resutls}.  

The main considerable point in the confusion matrix in Table \ref{tab:cm} is misclassification between categories E: esophagitis and I: normal-z-line. A large number of misclassifications like 30 images from the validation set occurred and a manual investigation was done to identify the reason. We notice that the images of these two categories were very similar to each other because of the close location in the GI tract, and identifying these is also a challeng for physicians.

\begin{table}
\centering
\tiny
  \caption{The Confusion Matrix of \textbf{Method 5} in our study} \vspace{-10pt}
  \caption*{\footnotesize \textmd{\textbf{A}:blurry-nothing, \textbf{B}:colon-clear, \textbf{C}:dyed-lifted-polyps, \textbf{D}:dyed-resection-margins, 
  \textbf{E}:esophagitis,\textbf{F}:instruments, \textbf{G}:normal-cecum, \textbf{H}:normal-pylorus, 
  \textbf{I}:normal-z-line, \textbf{J}:out-of-patient, \textbf{K}:polyps, \textbf{L}:retroflex-rectum, 
  \textbf{M}:retroflex-stomach, \textbf{N}:stool-inclusions, \textbf{O}:stool-plenty, \textbf{P}:ulcerative-colitis}}
  \vspace{-12pt}
  \label{tab:cm}
  \begin{tabular}{P{0.4mm}P{0.4mm}|P{0.4mm}|P{0.4mm}|P{1mm}|P{1mm}|P{1mm}|P{0.5mm}|P{1mm}|P{1mm}|P{1mm}|P{0.5mm}|P{1mm}|P{0.5mm}|P{1mm}|P{0.5mm}|P{1mm}|P{1mm}|}
                    \toprule
                    \multicolumn{18}{c}{Predicted class}   \\
                 
                    &   & \textbf{A}  & \textbf{B}  & \textbf{C}   & \textbf{D}   & \textbf{E}   & \textbf{F}  & \textbf{G}   & \textbf{H}   & \textbf{I}   & \textbf{J} & \textbf{K}   & \textbf{L}  & \textbf{M}   & \textbf{N}  & \textbf{O}   & \textbf{P}   \\
                    \midrule 
\parbox[t]{1mm}{\multirow{18}{*}{\rotatebox[origin=c]{90}{Actual class}}} & \textbf{A} & \textbf{53} & \_  & \_   & \_   & \_   & \_  & \_   & \_   & \_   & \_ & \_   & \_  & \_   & \_  & \_   & \_   \\
                    & \textbf{B} & \_  & \textbf{81} & \_   & \_   & \_   & \_  & \_   & \_   & \_   & \_ & \_   & \_  & \_   & \_  & \_   & \_   \\
                    & \textbf{C} & \_  & \_  & \textbf{130} & 7   & \_   & \_  & \_   & \_   & \_   & \_ & \_   & \_  & \_   & \_  & \_   & 1   \\
                    & \textbf{D} & \_  & \_  & 3   & \textbf{122} & \_   & \_  & \_   & \_   & \_   & \_ & \_   & \_  & \_   & \_  & \_   & \_   \\
                    & \textbf{E} & \_  & \_  & \_   & \_   & \textbf{115} & \_  & \_   & \_   & 19  & \_ & \_   & \_  & \_   & \_  & \_   & \_   \\
                    & \textbf{F} & \_  & \_  & \_   & \_   & \_   & \textbf{10} & \_   & \_   & \_   & \_ & 1   & \_  & \_   & \_  & \_   & \_   \\
                    & \textbf{G} & \_  & \_  & \_   & \_   & \_   & \_  & \textbf{125} & \_   & \_   & \_ & \_   & \_  & \_   & \_  & \_   & \_   \\
                    & \textbf{H} & \_  & \_  & \_   & \_   & \_   & \_  & \_   & \textbf{132} & \_   & \_ & \_   & \_  & \_   & \_  & \_   & \_   \\
                    & \textbf{I} & \_  & \_  & \_   & \_   & 11  & \_  & \_   & \_   & \textbf{121} & \_ & \_   & \_  & \_   & \_  & \_   & \_   \\
                    & \textbf{J} & \_  & \_  & \_   & \_   & \_   & 1  & \_   & \_   & \_   & \textbf{3} & \_   & \_  & \_   & \_  & \_   & \_   \\
                    & \textbf{K} & \_  & 1  & \_   & \_   & \_   & \_  & 6   & 2   & \_   & \_ & \textbf{172} & \_  & \_   & \_  & \_   & \_   \\
                    & \textbf{L} & \_  & \_  & \_   & \_   & \_   & \_  & 1   & \_   & \_   & \_ & \_   & \textbf{71} & \_   & \_  & \_   & \_   \\
                    & \textbf{M} & \_  & \_  & \_   & \_   & \_   & \_  & \_   & \_   & \_   & \_ & \_   & 2  & \textbf{118} & \_  & \_   & \_   \\
                    & \textbf{N} & \_  & \_  & \_   & \_   & \_   & \_  & \_   & \_   & \_   & \_ & \_   & \_  & \_   & \textbf{39} & \_   & \_   \\
                    & \textbf{O} & \_  & \_  & \_   & \_   & \_   & \_  & \_   & \_   & \_   & \_ & \_   & \_  & \_   & \_  & \textbf{110} & \_   \\
                    & \textbf{P} & \_  & \_  & \_   & \_   & 1   & 1  & 2   & \_   & \_   & \_ & 4   & 1  & \_   & \_  & \_   & \textbf{129} \\
                \bottomrule
\end{tabular}
\end{table}

\begin{table}
\small
  \caption{Validation results}
  \vspace{-12pt}
  \label{tab:our_results}
  \begin{tabular}{cccccccc}
    \toprule
    Method&REC&PREC&SPEC&ACC&MCC&F1&FPS\\
    \midrule
    1 & 0.855 & 0.793 & 0.989 & 0.816 & 0.814 & 0.823 & 79\\
    2 & 0.816 & 0.817 & 0.984 & 0.816 & 0.800 & 0.815 & 12\\
    3 & 0.9536 & 0.9543 & 0.9968 & 0.9536 & 0.9498 & 0.9535 & 64\\
    4 & 0.9555 & 0.9563 & 0.9969 & 0.9555 & 0.9519 & 0.9554 & 29\\
    5 & 0.9580 & 0.9587 & 0.9971 & 0.9580 & 0.9546 & 0.9580 & 29\\
  \bottomrule
\end{tabular}
\end{table}

\begin{table} 
\small 
  \caption{Official results}
\vspace{-10pt}
  \label{tab:org_resutls}
  \begin{tabular}{cccccccc}
    \toprule
    Method&REC&PREC&SPEC&ACC&MCC&F1\\
    \midrule
    1 & 0.8457 & 0.8457 & 0.9897 & 0.9807 & 0.8353 & 0.8456 \\
    2 & 0.8457 & 0.8457 & 0.9897 & 0.9807 & 0.8350 & 0.8457 \\
    3 & 0.9376 & 0.9376 & 0.9958 & 0.9922 & 0.9335 & 0.9376 \\
    4 & 0.9400 & 0.9400 & 0.9960 & 0.9925 & 0.9360 & 0.9400 \\
    5 & 0.9458 & 0.9458 & 0.9964 & 0.9932 & 0.9421 & 0.9458 \\
  \bottomrule
\end{tabular}
\vspace{-13pt}
\end{table}


\section{Conclusion}

In this paper, we presented five different methods for the multi-class classification of GI tract diseases. The proposed approach are based on the GFs, and pre-trained CNN with transfer learning mechanism. The combination of Resnet-152 and Densenet-161 with an additional MLP achieved the highest performance with both the validation dataset and the test dataset provided by the task organizers. We show that a combination of pre-trained deep neural models on ImageNet has better capabilities to classify images into the correct classes because of cumulative decision-making capabilities. For future work, we will combine deeper CNNs parallelly to add more cumulative decision taking capabilities for classifying multi-class objects. In addition to that, Generative Adversarial Network (GAN) methods can be utilized to handle imbalance dataset by generating more data to train deep neural networks.  


%
%


\clearpage

\bibliographystyle{ACM-Reference-Format}
\def\bibfont{\small} 
\balance
\bibliography{sigproc} 

\end{document}